\documentclass[sigconf]{acmart}

\setcopyright{none}
\renewcommand\footnotetextcopyrightpermission[1]{}

\usepackage{booktabs}   % professional tables
\usepackage{graphicx}   % figures
\usepackage{amsmath}    % math
\usepackage{hyperref}   % clickable links
\usepackage[ruled,vlined,linesnumbered]{algorithm2e} % pseudocode
\usepackage{tikz}           % diagrams
\usetikzlibrary{arrows.meta, positioning, calc}
\usepackage{pgfplots}       % bar charts
\pgfplotsset{compat=1.18}
\title{Proximity Features: Privacy-Compliant Cold-Start Personalization at Airbnb}

\author{Wei Jiang}
\affiliation{\institution{Airbnb Inc.}\city{San Francisco}\country{USA}}
\email{wei.jiang@airbnb.com}

\author{Bin Xu}
\affiliation{\institution{Airbnb Inc.}\city{San Francisco}\country{USA}}
\email{b.xu@airbnb.com}

\author{Hui Gao}
\affiliation{\institution{Airbnb Inc.}\city{San Francisco}\country{USA}}
\email{hui.gao@airbnb.com}

\author{Bharathi Thangamani}
\affiliation{\institution{Airbnb Inc.}\city{San Francisco}\country{USA}}
\email{bharathi.thangamani@airbnb.com}

\author{Weiwei Guo}
\affiliation{\institution{Airbnb Inc.}\city{San Francisco}\country{USA}}
\email{weiwei.guo@airbnb.com}

\author{Sundar Srinivasavaradhan}
\affiliation{\institution{Airbnb Inc.}\city{San Francisco}\country{USA}}
\email{sundara.srinivasavaradhan@airbnb.com}

\author{Tracy Yu}
\affiliation{\institution{Airbnb Inc.}\city{San Francisco}\country{USA}}
\email{tracy.yu@airbnb.com}

\author{Huiji Gao}
\affiliation{\institution{Airbnb Inc.}\city{San Francisco}\country{USA}}
\email{huiji.gao@airbnb.com}

\author{Michael Kinoti}
\affiliation{\institution{Airbnb Inc.}\city{San Francisco}\country{USA}}
\email{michael.kinoti@airbnb.com}

\acmConference[TSMO '26]{KDD Workshop on Two-sided Marketplace Optimization}{August 9, 2026}{Jeju, South Korea}



\begin{document}

\renewcommand{\shortauthors}{Jiang et al.}

\begin{abstract}
% Target: ~150 words
Personalization in two-sided marketplaces relies heavily on user-level features,
yet for platforms with infrequent, high-consideration purchases, a large fraction
of users lack sufficient history for effective recommendation, spanning
both paid and organic channels.
At Airbnb, a substantial share of search requests comes from logged-out or
first-time users, with this challenge especially pronounced on paid-channel
landing pages, leaving traditional user-level features unavailable for a large
fraction of traffic. Privacy regulations and
increasing restrictions on third-party cookies further limit identifier-based
tracking for non-essential use cases.
This paper introduces \emph{Proximity Features}, a privacy-compliant feature
system that groups users by geographic proximity using geo-IP data and an
adaptive clustering algorithm, producing aggregated user-level signals for
groups of approximately 1{,}000 nearby users---without requiring a persistent
individual identifier at inference time. Privacy is preserved by design: the
pipeline operates on consented, aggregated data only within consent-gated
privacy controls.

The system is deployed in production at Airbnb, serving multiple surfaces
including marketing landing pages and destination recommendation, with
engagement emails integration under way. Online A/B experiments demonstrate statistically significant
lifts in bookings, with the largest gains observed
among users with absent or stale history.

\end{abstract}

\maketitle

\section{Introduction}

Two-sided marketplaces such as Airbnb depend on personalization to match supply
with demand effectively. Machine learning models powering search ranking,
recommendation, and marketing rely on rich user-level features tied to
individual identifiers. However, a substantial portion of marketplace traffic
lacks the history needed to compute these features---a challenge known as the
\emph{cold-start problem}. Throughout this paper, \emph{user} is used as a
general term encompassing logged-in, logged-out, anonymous users, etc.

At Airbnb, the cold-start problem is pervasive across both paid and organic
channels. On paid-channel landing pages (e.g., Google, Meta), the majority of
users are logged out and many are first-time users with no prior Airbnb history.
Standard \texttt{user\_id} and cookie-based \texttt{visitor\_id}
features are either unavailable (for logged-out users) or too sparse to be useful (for first-time or low-activity users). Throughout this paper, \emph{cold-start} refers to both cases: users for whom individual identifiers are inaccessible and users with insufficient behavioral history.

Compounding this challenge, evolving privacy regulations impose additional
constraints. Under the General Data Protection Regulation
(GDPR)~\cite{voigt2017eu}, cookie-tracking requires explicit consent for
non-essential use cases such as marketing personalization. With increasing
restrictions on third-party cookies, identifier-based approaches face further
erosion for anonymous users. Any feature system targeting cold-start users in
this environment must work \emph{without individual tracking}.

This paper addresses both challenges simultaneously with \emph{Proximity
Features}, a feature system that leverages aggregated geo-IP signals from a
local group of nearby users. The key insight is that users from the same
geographic area exhibit correlated travel preferences---they tend to search for
and book similar destinations, prefer similar price ranges, and share
demographic characteristics. By grouping approximately 1{,}000 users into a
\emph{proximity bucket} using an adaptive clustering algorithm over geo-IP
coordinates, Proximity Features constructs rich user-level features immediately
available for any user whose IP address can be geolocated---regardless of login
state or prior history. Crucially, the system requires no user-level identifier
at inference time, making it applicable to anonymous users for whom traditional
personalization is otherwise unavailable.

The contributions of this paper are as follows:
\begin{enumerate}
  \item \textbf{New features via proximity key.} Proximity Features introduces
    a \emph{proximity key}---a compact identifier for a local geographic group
    of $\sim$1{,}000 users---that serves as a privacy-safe analog to
    \texttt{user\_id} in ML feature pipelines, enabling a new class of
    aggregated user-level features accessible across search ranking,
    recommendation, and marketing surfaces without requiring a persistent
    individual identifier at inference time.
  \item \textbf{Cold-start resolution via group aggregation.} An adaptive
    geo-IP clustering algorithm constructs proximity buckets at fine granularity
    in dense areas and coarser granularity in sparse areas, aggregating
    user-level signals from nearby users to fill the feature gap for users with
    absent or stale history.
  \item \textbf{Privacy-by-design without ID-based tracking.} The system
    operates entirely on aggregated, consented geo-IP data. No \texttt{user\_id}
    or cookie-based \texttt{visitor\_id} is required at inference time,
    making it resilient to third-party cookie deprecation and designed to
    operate within consent-gated, aggregated processing and deletion controls.
\end{enumerate}
The paper presents the core clustering algorithm and production-scale system
design (Section~\ref{sec:system}), followed by empirical results from A/B
experiments across multiple surfaces (Section~\ref{sec:experiments}).

\section{System Design}
\label{sec:system}

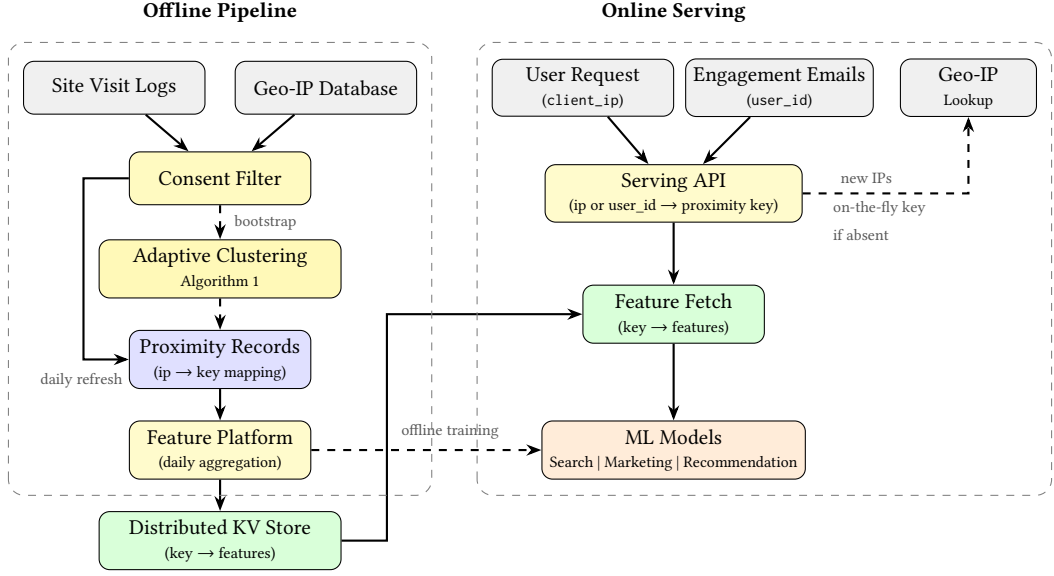
\begin{figure*}[t]
  \centering
  \begin{tikzpicture}[
  node distance=0.9cm and 1.2cm,
  box/.style={draw, rounded corners, minimum width=2.4cm, minimum height=0.7cm,
              align=center, font=\small},
  data/.style={box, fill=blue!12},
  process/.style={box, fill=yellow!25},
  store/.style={box, fill=green!15},
  model/.style={box, fill=orange!15},
  extdata/.style={box, fill=gray!12},
  arr/.style={-{Stealth[length=2mm]}, thick},
  darr/.style={-{Stealth[length=2mm]}, thick, dashed},
  label/.style={font=\small\bfseries, fill=white, inner sep=2pt},
  note/.style={font=\scriptsize, text=gray!70!black},
]

% ===== OFFLINE (left) =====
\node[label] (offline-label) at (-2.8, 5.8) {Offline Pipeline};

\node[extdata] (visits) at (-4.2, 4.8) {Site Visit Logs};
\node[extdata] (geoip) at (-1.4, 4.8) {Geo-IP Database};

\node[process, fill=yellow!30] (consent) at (-2.8, 3.6) {Consent Filter};

\node[process, fill=yellow!35, minimum width=3.2cm] (cluster) at (-2.8, 2.4)
  {Adaptive Clustering\\[-1pt]\scriptsize Algorithm~\ref{alg:clustering}};

\node[data] (records) at (-2.8, 1.2) {Proximity Records\\[-1pt]\scriptsize(ip $\to$ key mapping)};

\node[process] (featcomp) at (-2.8, 0.0) {Feature Platform\\[-1pt]\scriptsize(daily aggregation)};

\node[store, minimum width=3.2cm] (kvstore) at (-2.8, -1.2)
  {Distributed KV Store\\[-1pt]\scriptsize(key $\to$ features)};

% Offline arrows
\draw[arr] (visits) -- (consent);
\draw[arr] (geoip) -- (consent);
% Bootstrap path (dashed): consent -> clustering -> records
\draw[darr] (consent) -- (cluster) node[note, midway, right, xshift=2pt] {bootstrap};
\draw[darr] (cluster) -- (records);
% Daily path (solid): consent -> records (routed around left side)
\draw[arr] (consent.west) -- ++(-0.6,0) |- (records.west)
  node[note, pos=0.5, left, xshift=18pt, yshift=-8pt] {daily refresh};
% Records onward
\draw[arr] (records) -- (featcomp);
\draw[arr] (featcomp) -- (kvstore);

% ===== ONLINE (right) =====
\node[label] (online-label) at (3.2, 5.8) {Online Serving};

% Two input sources side by side
\node[extdata] (visitor) at (2.0, 4.8) {User Request\\[-1pt]\scriptsize(\texttt{client\_ip})};
\node[extdata] (emails) at (4.6, 4.8) {Engagement Emails\\[-1pt]\scriptsize(\texttt{user\_id})};

\node[process, minimum width=3.4cm] (api) at (3.2, 3.4)
  {Serving API\\[-1pt]\scriptsize(ip or user\_id $\to$ proximity key)};

\node[store] (kvonline) at (3.2, 1.8)
  {Feature Fetch\\[-1pt]\scriptsize(key $\to$ features)};

\node[model, minimum width=3.4cm] (models) at (3.2, 0.0)
  {ML Models\\[-1pt]\scriptsize Search \textbar{} Marketing \textbar{} Recommendation};

% Online arrows
\draw[arr] (visitor) -- (api);
\draw[arr] (emails) -- (api);
\draw[arr] (api) -- (kvonline);
\draw[arr] (kvonline) -- (models);

% KV store to online fetch (cross-connection)
\draw[arr] (kvstore.east) -- ++(0.6,0) |- (kvonline.west);

% Feature Platform to ML models (offline training)
\draw[darr] (featcomp.east) -- ++(0.6,0) |- (models.west);
\node[note, anchor=east] at (1.0, 0.25) {offline training};

% On-the-fly key computation annotation
\node[note, anchor=west] at (5.2, 3.2) {on-the-fly key};
\node[note, anchor=west] at (5.2, 2.85) {if absent};

% Geo-IP fallback for new IPs
\node[extdata, minimum width=1.8cm] (geoonline) at (7.1, 4.8)
  {Geo-IP\\[-1pt]\scriptsize Lookup};
\draw[darr] (api.east) -| (geoonline.south);
\node[note, anchor=east] at (6.2, 3.6) {\scriptsize new IPs};

% Dashed box around offline
\draw[dashed, gray, rounded corners=6pt]
  (-5.6, -0.6) rectangle (-0.0, 5.4);

% Dashed box around online
\draw[dashed, gray, rounded corners=6pt]
  (0.6, -0.6) rectangle (8.2, 5.4);

\end{tikzpicture}
  \caption{System architecture. Offline: the adaptive clustering
    algorithm runs once at bootstrap to establish the proximity key partition;
    the IP-to-key mapping, coordinate-to-scalar mapping, and tile-to-bucket-count
    mapping are refreshed daily, and features are aggregated over
    stable keys and stored in a distributed KV store. Online: a
    user's IP is mapped to a proximity key via a serving API (with on-the-fly
    computation for unseen IPs), and features are fetched for model scoring.
    Engagement emails pass a user ID to the same API, which internally
    resolves it to an IP.}
  \Description{System architecture diagram showing the offline pipeline (geo-IP clustering, feature computation, KV store) and online serving path (IP to proximity key lookup, feature fetch, model scoring).}
  \label{fig:architecture}
\end{figure*}

Figure~\ref{fig:architecture} provides an overview of the Proximity Features
system. Each component is described below.

\subsection{Geo-IP Data Source}

Geo-IP databases~\cite{poese2011ip} map IP addresses to
latitude/longitude coordinates. A regularly updated geo-IP data source covers
the vast majority of Airbnb's global traffic. In our internal analysis,
city-level agreement is approximately 80\%. Most IP addresses map to a single,
stable latitude/longitude pair, while approximately 15\% map to different
coordinates on different days due to ISP address reassignment.

Importantly, coordinates from the geo-IP database are aggregated around the center of population
rather than precise street-level locations---they cannot be used to identify
specific addresses or households, even though they include more than two decimal
places of precision. As a result, multiple IP addresses naturally map to the
same latitude/longitude pair, typically representing a city center. This
property is foundational to the privacy design described in
Section~\ref{sec:privacy}.

As a data quality measure, bot traffic is filtered prior to clustering. Automated
crawlers can account for a disproportionate share of requests at certain IP
addresses, skewing user counts and distorting bucket sizes. Removing
known-bot traffic ensures that the adaptive clustering algorithm operates on
a representative distribution of human traffic.

\subsection{Proximity Key}

A \emph{proximity key} uniquely identifies a geographic bucket of users. It
is constructed from three components:
\begin{equation}
  \text{key} = \text{base64}\bigl(\lfloor \text{lat} \times s \rfloor,\;
    \lfloor \text{lng} \times s \rfloor,\;
    h(\text{ip}) \bmod b\bigr)
  \label{eq:key}
\end{equation}
where $s$ is the \emph{proximity scalar} controlling geographic tile size, $b$
is the number of IP hash buckets within that tile, and $h$ is a deterministic
hash function (\texttt{xxhash64}~\cite{xxhash} is used for consistency between offline and online paths). The key is base64-encoded for compact storage and transmission.

For dense locations (Phase~1 of the clustering algorithm), $s$ corresponds to
the native precision of the geo-IP coordinates and $b > 1$ provides
sub-location granularity via IP hashing. For sparse locations (Phase~2), $s$
is set to a coarser quantization factor and $b = 1$. The scalar $s$ and bucket
count $b$ are determined per tile by the adaptive clustering algorithm described
next.

The proximity key serves as a privacy-safe analog to \texttt{user\_id}: it
identifies a local geographic area rather than an individual, is small and
precise for densely populated areas, and coarser for sparse regions. For
example, in a dense urban area, an IP address maps to a key encoding a
fine-grained tile with multiple IP hash buckets:
{\small\texttt{base64("lat:4071,lng:-7401,bucket:2")}}.
In a sparse rural area, the same structure encodes a coarser tile with a
single bucket:
{\small\texttt{base64("lat:39,lng:-105,bucket:0")}}.
Any ML model that currently conditions on \texttt{user\_id} can condition on
the proximity key instead to serve anonymous and cold-start users.

\subsection{Adaptive Clustering Algorithm}
\label{sec:clustering}

Proximity bucketing is motivated by the following \emph{constrained spatial
partitioning} objective. Let $\mathcal{L} = \{(l_i, g_i, n_i)\}_{i=1}^{N}$ be
the set of $N$
distinct latitude/longitude pairs with their associated user counts. A spatially
contiguous partition $\mathcal{P} = \{B_1, \ldots, B_m\}$ of $\mathcal{L}$ is
sought that maximizes the number of buckets $m$---preserving the finest
geographic granularity---subject to a minimum bucket size constraint:
\begin{equation}
  \max_{|\mathcal{P}|} \quad \text{s.t.} \quad
  \sum_{i : (l_i, g_i, n_i) \in B_j} n_i \;\geq\; T \quad \forall\, B_j \in \mathcal{P}
  \label{eq:partition}
\end{equation}
where $T$ is the target bucket size. The spatial contiguity requirement ensures
that each bucket corresponds to a geographic neighborhood; maximizing $m$
preserves the finest granularity possible. The threshold $T \approx 1{,}000$ ensures each bucket
aggregates over sufficiently many users, serving a dual objective: it
is intended to provide a $k$-anonymity-like privacy
property~\cite{sweeney2002kanonymity}, and it provides enough observations
for stable estimation of user-level features (e.g., destination distributions).
Because the algorithm is greedy, $T$ is a target rather
than a strict guarantee---some buckets may be smaller, particularly in
low-traffic regions assigned at the coarsest resolution.

The challenge is that user density varies by orders of magnitude: a single
coordinate near a major airport may correspond to tens of thousands of users,
while rural coordinates may have only a handful. Solving
Equation~\ref{eq:partition} exactly is NP-hard~\cite{garey1979computers}; it is an instance of
the bin covering problem~\cite{assmann1984dual}. Rather than pursuing
theoretical approximation guarantees, a greedy two-phase heuristic is proposed
that exploits the structure of geo-IP data for computational efficiency and
practical effectiveness (Algorithm~\ref{alg:clustering}).

\begin{algorithm*}[t]
\caption{Adaptive Proximity Clustering}
\label{alg:clustering}
\KwIn{Set of (lat, lng, ip) tuples from consented site visits; target bucket size $T$; minimum IP count $n_{\text{ip}}^{\min}$; scalar range $[e_{\min}, e_{\max}]$; ratio threshold $\rho$}
\KwOut{Proximity key assignment for each (lat, lng) pair}

\BlankLine
\tcp{Phase 1: Dense locations --- zoom in via IP hash buckets}
\ForEach{unique (lat, lng) pair}{
  $n_v \leftarrow$ number of unique users at (lat, lng)\;
  $n_{\text{ip}} \leftarrow$ number of unique IP addresses at (lat, lng)\;
  \If{$n_v \geq T$ \textbf{and} $n_{\text{ip}} \geq n_{\text{ip}}^{\min}$}{
    $b \leftarrow \lceil n_v / T \rceil$\;
    Assign each user to bucket $({\text{lat}},\; {\text{lng}},\; h(\text{ip}) \bmod b)$\;
    Mark (lat, lng) as \emph{assigned}\;
  }
}

\BlankLine
\tcp{Phase 2: Sparse locations --- zoom out via multi-pass coarsening}
$R \leftarrow$ all unassigned (lat, lng) pairs\;
\For{$e \leftarrow e_{\max}$ \KwTo $e_{\min}$}{
  $s \leftarrow 10^{e}$\;
  Quantize each $(lat, lng) \in R$ to tile $(\lfloor lat \cdot s \rfloor,\; \lfloor lng \cdot s \rfloor)$\;
  Compute $n_v$ per quantized tile\;
  $r \leftarrow$ fraction of tiles with $n_v \geq T$\;
  \If{$r \geq \rho$ \textbf{or} $e = e_{\min}$}{
    \ForEach{tile with $n_v \geq T$ \textbf{(}or all remaining tiles if $e = e_{\min}$\textbf{)}}{
      Assign all (lat, lng) in tile with scalar $s$ and $b = 1$\;
    }
    $R \leftarrow R \setminus \text{assigned pairs}$\;
    \lIf{$R = \emptyset$}{\textbf{break}}
  }
}
\end{algorithm*}

\paragraph{Phase 1: Dense locations (refinement).}
For latitude/longitude pairs with $\geq T$ unique users \emph{and}
$\geq n_{\text{ip}}^{\min}$ unique IP addresses, the location is \emph{refined}
by subdividing using IP hash buckets. The number of buckets is set to
$b = \lceil n_{\text{users}} / T \rceil$, targeting buckets of
approximately $T$ users each. These locations retain the original
(full-precision) latitude/longitude as their tile coordinates, with the IP
bucket index providing sub-tile granularity. This phase maximizes geographic granularity for dense locations by retaining
native coordinate precision; the target of approximately $T$ users per bucket
is best-effort, as hash-based assignment does not guarantee perfectly uniform
distribution.

\paragraph{Phase 2: Sparse locations (coarsening).}
For the remaining locations, a multi-pass coarsening procedure is applied,
analogous to adaptive mesh refinement~\cite{berger1984adaptive}, where
resolution is reduced until sufficient mass accumulates in each cell. The
algorithm iterates over decreasing scalar values $s = 10^e$ for
$e \in \{e_{\max}, e_{\max}-1, \ldots, e_{\min}\}$. At each pass, all remaining coordinates are quantized by $s$ and the number of
unique users per quantized tile is computed. When the fraction of qualifying tiles (those with
$n_v \geq T$) meets the ratio threshold $\rho$, they are \emph{assigned} at
this scalar level, and their constituent coordinates are removed from
subsequent passes. This continues until all coordinates are assigned or the
minimum scalar $e_{\min}$ is reached.

The result is an adaptive resolution map: urban centers receive fine-grained
tiles (high $s$), while rural areas are grouped into coarser tiles (low $s$).
No IP bucketing is applied for sparse locations ($b = 1$). The greedy
fine-to-coarse strategy ensures that each assigned tile satisfies the bucket
size constraint at the finest resolution possible, capturing the same design
tradeoff as Equation~\ref{eq:partition}: maximizing geographic granularity
subject to bucket-size constraints.

\paragraph{Complexity.}
Each pass requires a single group-by aggregation over the remaining data. With
$k$ passes ($k \leq e_{\max} - e_{\min} + 1$, typically $\leq 10$), the total
cost is $O(kN)$ where $N$ is the number of distinct latitude/longitude pairs.
Because coordinates from the geo-IP database are pre-aggregated around population centers, $N$ is
orders of magnitude smaller than the number of raw IP addresses---substantially
cheaper than general spatial clustering
algorithms~\cite{ester1996density, bentley1975multidimensional}.

\paragraph{Parameter selection.}
The algorithm is governed by a small set of parameters: target bucket size $T$,
minimum IP count $n_{\text{ip}}^{\min}$, scalar range $[e_{\min}, e_{\max}]$,
and ratio threshold $\rho$. Exact production settings are omitted because they
were tuned on proprietary traffic distributions and are tightly coupled to
Airbnb-specific site visit volume. Illustrative values such as
$T \approx 1{,}000$ and $n_{\text{ip}}^{\min} = 100$ are provided only to
convey the operating regime, not as universal or production defaults. A grid of
candidate values was evaluated using offline destination prediction accuracy
(Recall@$k$), selecting the combination that maximized predictive signal while
maintaining the bucket size target. This empirical tuning is part of the greedy
design: rather than solving the partitioning problem exactly, the heuristic is
parameterized and values are selected that perform best in practice.

\paragraph{Key stability.}
Proximity records are stable at multiple levels. Latitude/longitude coordinates
from the geo-IP database are inherently stable, as they are anchored to
population centers rather than individual addresses. The IP-to-coordinate
mapping is also mostly stable: the majority of IP addresses resolve to the same
coordinates over time, with only a small fraction shifting due to ISP address
reassignment. Together, these properties mean that proximity keys change
infrequently---the same key assignment can serve production traffic for years
without re-clustering. The key partition used throughout this paper was
bootstrapped on 2023 site visit data and has remained in production without
re-clustering since. The daily batch pipeline handles incremental changes
(new IPs, occasional coordinate shifts, etc.) by refreshing the IP-to-key mapping,
the coordinate-to-scalar mapping, and the tile-to-bucket-count mapping.
Features, by contrast, are refreshed daily, reflecting evolving user signals
over the stable key partition. This decoupling of keys and features minimizes
maintenance overhead while keeping signals fresh.

Proximity keys are also robust to IP address reassignment. Since proximity
features represent aggregate group signals rather than individual behavior,
geographic consistency at the group level is what matters---not per-IP
stability. An IP that shifts from a suburban area to a nearby city center is
reassigned to the corresponding proximity bucket; its signals enrich
both buckets over time. The only failure case would be a gross misassignment
(e.g., an IP in Los Angeles resolving to New York), which is rare and mitigated
by the geo-IP data quality checks described above.

\subsection{Feature Computation}

Features are computed daily over each proximity bucket and stored in a
distributed key-value store. They fall into three categories:
\textbf{short-term engagement} (7/28/56-day windows: top destinations, room
types, median price, search parameters);
\textbf{long-term booking patterns} (28/90/365-day windows: booked
destinations, listing attributes, travel party characteristics); and
\textbf{demographics} (birth decade distribution, bucket size, geographic
metadata).

Short-term features additionally capture search intent signals: typical values
for party size, trip duration, listing prices, and travel distance. Long-term
features include travel party attributes such as the fraction of bookings
involving pets, families, or groups---signals otherwise unavailable for
cold-start users. Multiple aggregation windows are retained rather than
collapsed into a single value, allowing downstream models to select the time
horizon most predictive for their task, as validated by the offline recall
results in Section~\ref{sec:experiments}.

\subsection{Serving}

Proximity features support both online and offline consumption via a unified
serving API. For real-time surfaces, the \texttt{client\_ip} of the visiting
user, as sourced from the CDN, is resolved to a proximity key---first via a
daily batch mapping, falling back to on-the-fly
key computation for unseen IPs---and the corresponding features are fetched from
the KV store. The API is a \emph{soft dependency}: models proceed with a
degraded feature vector if the lookup exceeds a 60ms timeout. For batch channels
(e.g., engagement emails), callers pass a \texttt{user\_id}; the API resolves it
to a recent IP and follows the same flow.

For unseen IPs, the on-the-fly algorithm computes a proximity key in real time
without requiring a full re-clustering: (1) the IP is geolocated to
(lat,~lng); (2) the proximity scalar $s$ for that coordinate is fetched
from the KV store; (3) if no scalar exists, a default of $s{=}1$ is
used; (4) the quantized geo tile is computed; (5) the bucket count $b$ for that
tile is fetched from the KV store; and (6) the key is computed as:
\[
  \text{base64}\bigl(\lfloor\text{lat}{\cdot}s\rfloor,\;\lfloor\text{lng}{\cdot}s\rfloor,\;h(\text{ip})\bmod b\bigr).
\]
This mirrors the offline key construction (Equation~\ref{eq:key}) exactly,
ensuring new IPs always receive a valid, consistent proximity key. If the bucket count lookup misses, the API returns no proximity key and the
model proceeds via the soft dependency mechanism; this occurs in fewer than
0.5\% of requests in practice.

\subsection{Privacy Design}
\label{sec:privacy}

Proximity Features achieve privacy through layered obfuscation, beginning with
data selection. The input dataset for the clustering algorithm excludes users
who did not provide consent, respecting user choice from the outset.

At the geolocation layer, the latitude and longitude coordinates from the geo-IP database
are already aggregated around population centers---they cannot identify
specific addresses or households, even though they include more than two decimal
places of precision. As a result, multiple IP addresses naturally map to the
same latitude/longitude pair, typically representing a city center.

Note that users connecting via VPN are assigned to the proximity bucket of the
VPN exit node rather than their physical location; this reflects a deliberate
user choice and is consistent with the system's privacy-respecting design.

Building on this inherent coarseness, the adaptive clustering algorithm applies
additional aggregation to form local groups targeting approximately 1{,}000 users.
All modeling is performed at this group level: the features
associated with a proximity key reflect the collective patterns of the group,
not the actions of any individual. This ensures that individual user behavior
is masked within the aggregate, avoiding individual tracking entirely.

Beyond the algorithmic design, the system enforces operational safeguards:
\begin{itemize}
  \item \textbf{No persistent individual identifier at inference.} The
    real-time serving path requires only \texttt{client\_ip}---no persistent
    user ID or cookie-based ID is needed. (Offline callers may pass a
    \texttt{user\_id}, which the API resolves to an IP internally.)
  \item \textbf{Consent-gated pipeline.} Both the clustering input data and the
    feature computation pipeline are integrated with a consent management
    framework, ensuring only consented data is used throughout.
  \item \textbf{Privacy and data-governance controls.} PII tables have short
    retention windows, and non-PII tables undergo regular selective deletion in
    accordance with data governance policies.
\end{itemize}

\section{Applications}
\label{sec:applications}

Proximity Features are designed as a reusable feature layer that can be
integrated into any ML model receiving user traffic. Two launched applications
and one offline use case are described below.

\subsection{Marketing Landing Pages}

The term Marketing Landing Pages refers to pages where users arrive from external
channels such as paid advertising (e.g., Google and Meta ads) and organic
search, landing on customized pages displaying personalized listing
recommendations. These pages have the highest cold-start rate across Airbnb
surfaces, and the listing recommender operated with severely sparse feature
vectors for the majority of traffic prior to proximity features.

By incorporating proximity features, the model gains access to the booking and
browsing patterns of nearby users, enabling meaningful personalization even
for completely anonymous traffic. This application was launched across two
marketing-entry surfaces in December 2024 and January 2025. For external
presentation, these two separate experiments are grouped under the label
\emph{Marketing Landing Pages} because they share a similar cold-start listing
recommendation setting. The A/B results in Table~\ref{tab:ab} report aggregate lift
across both surfaces.

\subsection{Destination Recommendation}

The Airbnb homepage features a destination suggestion module (AutoSuggest) that
recommends travel destinations as users interact with the search bar. This
module is powered by a destination intent model that predicts which destinations
a user is likely to book.

The baseline model (V1) relied on 15 features derived from user search and
booking history. For logged-out or first-time users, these features were
empty, and the model fell back to a static list of globally popular
destinations---the same recommendations regardless of the user's location or
context.

The updated model (V2) incorporates 26 features (13 new, 2 removed), including proximity
features such as the most-booked destinations and most-viewed locations by
nearby users. With these signals, the model replaces generic recommendations
with location-aware suggestions. For example, a user from Seoul now sees
destinations popular among nearby travelers (e.g., Jeju Island, Osaka) rather
than a global default list.

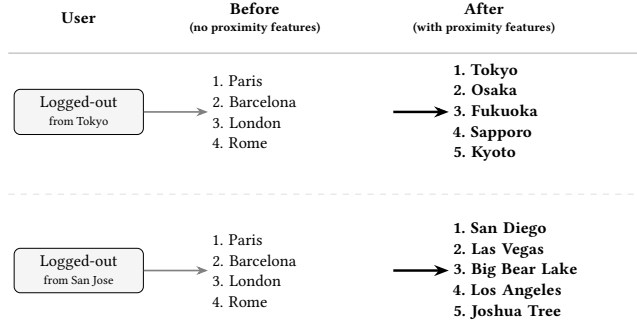
\begin{figure}[t]
  \centering
  \resizebox{\columnwidth}{!}{\begin{tikzpicture}[
  font=\small,
  userbox/.style={draw, rounded corners=3pt, fill=gray!8, minimum width=2.2cm,
    minimum height=0.7cm, align=center, font=\small},
  listbox/.style={align=left, font=\small, text width=3.0cm, inner sep=2pt},
  header/.style={font=\small\bfseries, align=center},
  arr/.style={-{Stealth[length=2mm]}, thick, gray},
  arrbold/.style={-{Stealth[length=2.5mm]}, very thick, black},
]

% Headers
\node[header] at (0, 2.8) {User};
\node[header] at (3.0, 2.8) {Before\\[-2pt]\scriptsize(no proximity features)};
\node[header] at (6.9, 2.8) {After\\[-2pt]\scriptsize(with proximity features)};

% Horizontal rule
\draw[gray!40, thick] (-1.2, 2.2) -- (9.6, 2.2);

% --- Row 1: Tokyo ---
\node[userbox] (u1) at (0, 1.2) {Logged-out\\[-1pt]\scriptsize from Tokyo};

\node[listbox, anchor=west] (b1) at (2.2, 1.2) {%
  1.~Paris\\
  2.~Barcelona\\
  3.~London\\
  4.~Rome};

\node[listbox, anchor=west, font=\small\bfseries] (a1) at (6.3, 1.2) {%
  1.~Tokyo\\
  2.~Osaka\\
  3.~Fukuoka\\
  4.~Sapporo\\
  5.~Kyoto};

\draw[arr] (u1.east) -- (b1.west);
\draw[arrbold] (b1.east) -- (a1.west);

% Divider
\draw[gray!25, dashed] (-1.2, -0.2) -- (9.6, -0.2);

% --- Row 2: San Jose ---
\node[userbox] (u2) at (0, -1.5) {Logged-out\\[-1pt]\scriptsize from San Jose};

\node[listbox, anchor=west] (b2) at (2.2, -1.5) {%
  1.~Paris\\
  2.~Barcelona\\
  3.~London\\
  4.~Rome};

\node[listbox, anchor=west, font=\small\bfseries] (a2) at (6.3, -1.5) {%
  1.~San Diego\\
  2.~Las Vegas\\
  3.~Big Bear Lake\\
  4.~Los Angeles\\
  5.~Joshua Tree};

\draw[arr] (u2.east) -- (b2.west);
\draw[arrbold] (b2.east) -- (a2.west);

\end{tikzpicture}}
  \caption{Qualitative comparison of destination suggestions for logged-out
    users. Without proximity features, the model returns a static global
    list; with them, suggestions reflect local travel patterns.}
  \Description{Side-by-side comparison of destination suggestions with and without proximity features for two logged-out users, showing location-aware recommendations replacing generic global defaults.}
  \label{fig:qualitative}
\end{figure}

Figure~\ref{fig:qualitative} illustrates this effect for two logged-out
users. Separately, architectural improvements made V2 37\% smaller than V1 in
parameter count, reducing inference latency; this is orthogonal to the
proximity feature contribution. This application was launched in
March 2026.

\subsection{Engagement Emails}

Beyond real-time surfaces, proximity features are being integrated into
offline batch use cases. In engagement email campaigns, Airbnb sends
personalized listing and destination recommendations to users. The serving
infrastructure is in place: each recipient's \texttt{user\_id} can be passed
to the same serving API, which internally resolves it to the most recently
logged IP address and retrieves the corresponding proximity key and features
(Section~\ref{sec:system}). Experiment integration is currently underway to
validate the lift for users with limited booking history.

\section{Experiments}
\label{sec:experiments}

Proximity Features are evaluated through both offline metrics and online A/B
experiments across production surfaces.

\subsection{Offline Evaluation}

To assess the predictive signal of proximity features in isolation, the top-$k$
destinations are extracted from proximity feature histograms (booked and viewed
geos at various time windows) and Recall@$k$ is measured against actual bookings
in a subsequent 90-day period.

\paragraph{Setup.} User--feature snapshots are sampled from a 90-day observation
period. Each feature has its own aggregation window (e.g., 7-day, 28-day, or
365-day), reflecting different time horizons. Booking labels are sourced from
the subsequent 90 days.
Results are reported for \emph{low-activity} users---those with limited booking
history---the primary target population for proximity features.

\begin{table}[t]
  \centering
  \caption{Recall@$k$ (\%) for destination prediction using proximity features
    on low-activity users across different aggregation windows. No comparative
    baseline is included; the evaluation isolates proximity feature signal in
    the cold-start setting.}
  \label{tab:recall}
  \begin{tabular}{lccc}
    \toprule
    Feature & R@1 & R@3 & R@5 \\
    \midrule
    Booked geos (365d) & 12.06 & 22.88 & 29.14 \\
    Booked geos (90d)  & 11.91 & 22.20 & 28.25 \\
    Viewed geos (28d)  & 11.90 & 22.96 & 29.31 \\
    Viewed geos (7d)   & 11.81 & 23.12 & 30.03 \\
    \bottomrule
  \end{tabular}
\end{table}

Table~\ref{tab:recall} shows that proximity features alone achieve Recall@5
of approximately 29--30\% for geo-level destination prediction among low-activity
users, indicating strong geographic signal. Notably, short-term viewed-geo features (7-day window)
perform comparably to long-term booked-geo features (365-day window),
suggesting that recent browsing activity within a proximity bucket carries
substantial predictive value.

In an offline ablation for the Destination Intent Model V2
(Section~\ref{sec:applications}), adding proximity features yielded an
approximately 2\% relative improvement in Top-3 accuracy.

\subsection{Online A/B Experiments}

Results are reported from production A/B experiments, each comparing a model
with proximity features (treatment) against the same model without them
(control). All experiments ran for sufficient duration to reach statistical
significance at the 95\% confidence level. Notably, the results below span
December 2024 through March 2026, all using the same proximity key assignment,
empirically validating the temporal stability of the key partition.

\begin{table}[t]
  \centering
  \caption{Online A/B experiment results across production surfaces.}
  \label{tab:ab}
  \begin{tabular}{llr}
    \toprule
    Surface & Metric & Lift \\
    \midrule
    Marketing Landing Pages & Global Bookings & +0.011\% \\
    Marketing Landing Pages & First-time Bookers & +2.0\% \\
    Marketing Landing Pages & Searchers w/ Dates & +0.054\% \\
    \midrule
    AutoSuggest & Global Bookings & +0.16\% \\
    AutoSuggest & First-time Bookings & +0.33\% \\
    \bottomrule
  \end{tabular}
\end{table}

Table~\ref{tab:ab} summarizes the results. On marketing landing pages, proximity features produced a +0.011\% lift in
global bookings and increased first-time bookers by +2.0\%, confirming
effectiveness for the cold-start population. Searchers with dates---a key
engagement metric indicating users progressing further in the booking
funnel---increased by +0.054\%.

On AutoSuggest, the Destination Intent Model V2 with proximity features
achieved a +0.16\% lift in global bookings and +0.33\% in first-time bookings. This represents the largest booking lift from proximity features across all
production applications to date.

\subsection{Analysis}

The following analysis focuses on the AutoSuggest experiment, which provides the
richest cohort-level breakdown. Three cohorts reflect booking history at
assignment time: \emph{Never-Booked} (no prior bookings, reported separately
from assignment-date New-Guest users; users may still book during the experiment), \emph{Active} (booked within the past year), and
\emph{Dormant} (booked previously but not within the past year).
\emph{New-Guest} is an assignment-date booking-event cohort for users who also
have no prior bookings and whose first-ever booking occurs on the assignment date.

\begin{figure}[t]
  \centering
  \begin{tikzpicture}
\begin{axis}[
  ybar,
  width=\columnwidth,
  height=4.5cm,
  bar width=14pt,
  ylabel={Booking Lift (\%)},
  ylabel style={font=\small},
  symbolic x coords={Never-Booked, Dormant, Active, New-Guest},
  xtick=data,
  xticklabel style={font=\small},
  yticklabel style={font=\small},
  ymin=-0.15, ymax=0.65,
  ytick={0, 0.1, 0.2, 0.3, 0.4, 0.5, 0.6},
  nodes near coords,
  every node near coord/.append style={
    font=\scriptsize\bfseries,
    /pgf/number format/fixed,
    /pgf/number format/precision=2,
    /pgf/number format/fixed zerofill,
    above,
  },
  axis lines*=left,
  enlarge x limits=0.2,
  ymajorgrids=true,
  xmajorgrids=false,
  major grid style={gray!20},
]
\addplot[
  fill=blue!50, draw=blue!70,
] coordinates {
  (Never-Booked, 0.33)
  (Dormant, 0.27)
  (Active, 0.15)
  (New-Guest, -0.04)
};
\end{axis}
\end{tikzpicture}

\smallskip
{\small
\begin{tabular}{lrr}
  \toprule
  Cohort & Traffic Share & P-Value \\
  \midrule
  Never-Booked     & 65\%  & 0.17 \\
  Dormant       & 11\%  & 0.11 \\
  Active        & 23\%  & 0.16 \\
  New-Guest & ${<}1\%$ & 0.78 \\
  \bottomrule
\end{tabular}
}
  \caption{Booking lift by user cohort in the AutoSuggest experiment, with
    traffic share and p-value. The directionally largest gains are among Never-Booked and
    Dormant users. The New-Guest cohort ($<$1\% of traffic,
    $p=0.78$) shows a near-zero non-significant result.}
  \Description{Bar chart showing booking lift by user cohort (Never-Booked, Dormant, Active, New-Guest) with a companion table showing traffic share and p-value for each cohort.}
  \label{fig:cohorts}
\end{figure}
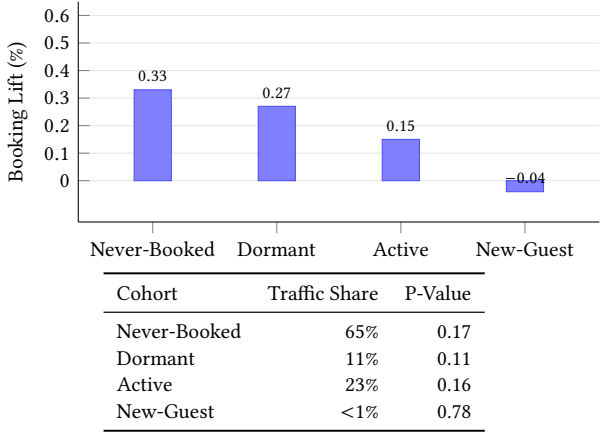

\paragraph{Cohort effects.}
The directionally largest booking lifts were observed among
\emph{Never-Booked} and \emph{Dormant} user cohorts (Figure~\ref{fig:cohorts}).
The three largest cohorts (Never-Booked, Active, Dormant) show consistent positive
directional lifts (+0.15--0.33\%) in global bookings, though individually below the 95\%
significance threshold due to limited subgroup power. Even Active users benefit directionally, as infrequent booking behavior keeps
individual history sparse. The New-Guest cohort
($<$1\% of experiment traffic, $-0.04\%$, $p=0.78$) is the smallest and shows a near-zero,
non-significant result, consistent with insufficient power rather than a true
negative effect. Separately, the +0.33\% first-time bookings lift (Table~\ref{tab:ab})
is an experiment-level outcome metric capturing first-ever bookings among
users with no prior bookings at assignment, whether reported as assignment-date
New-Guest users or as Never-Booked users who book later in the experiment window.
The gains are concentrated among users with absent or stale history---most
clearly Never-Booked and Dormant users---for whom the baseline model could only
serve a static list of popular destinations. Proximity features fill this gap by
providing location-aware signals derived from nearby users, enabling personalized
recommendations from the first interaction.
These results are consistent with the design hypothesis that proximity features provide the most incremental value where user-level features are absent or stale.

\paragraph{Impression distribution shift.}
The treatment arm also surfaced specific international
destinations (e.g., Kuala Lumpur, Seoul, Jeju Island, Dubai) that the baseline
model rarely recommended. This indicates that proximity features enable the
model to produce more diverse, location-relevant recommendations rather than
falling back to generic defaults. The shift also reflects a reduction in
popularity bias~\cite{abdollahpouri2019popularity}: rather than defaulting to
globally popular destinations, the model surfaces destinations that are locally
relevant to the user's geographic context. This is a direct consequence of
replacing a static global prior with group-level signals derived from nearby
users.

\section{Related Work}

\paragraph{Spatial indexing and geo bucketing.}
Geohash~\cite{niemeyer2008geohash}, Google S2~\cite{s2geometry}, and Uber
H3~\cite{brodsky2018h3} provide hierarchical spatial indexing at fixed
resolutions. While computationally efficient, fixed-resolution schemes cannot
adapt to the extreme variation in user density across geographies: a single
cell may contain millions of users in a city center and only a handful in a
rural area. KD-trees~\cite{bentley1975multidimensional} offer adaptive
partitioning but require $O(n \log n)$ construction and are designed for point
queries rather than group-level feature aggregation.
Quadtrees~\cite{finkel1974quad} recursively subdivide space into cells of
varying size, and the Phase~2 coarsening is similar in spirit---but operates on
a discrete set of power-of-ten scalars rather than recursive bisection. The
approach also draws on the principle of adaptive mesh
refinement~\cite{berger1984adaptive}---adjusting resolution to local
density---while operating on a discrete geographic grid with $O(kN)$ cost.

\paragraph{Cold-start recommendation.}
The cold-start problem is well-studied in recommender
systems~\cite{schein2002methods, lam2008addressing}. Common approaches include
content-based filtering~\cite{pazzani2007content}, hybrid
models~\cite{burke2002hybrid}, and demographic-based
approaches that leverage user attributes when behavioral data is
sparse. In the marketplace setting,
\citet{haldar2020improving} describe Airbnb's approach to improving deep
learning for search ranking, and \citet{grbovic2018real} address real-time
personalization using embeddings. Proximity Features are complementary: they provide a group-level signal layer
that can augment any of these approaches when individual-level data is
unavailable.

\paragraph{Privacy-preserving ML}
Federated learning~\cite{mcmahan2017communication} and differential
privacy~\cite{dwork2006calibrating} enable model training without exposing
individual data. Proximity Features take a different approach: rather than modifying the training
procedure, features are constructed over aggregated groups so that individual
behavior is masked by design. As discussed in Section~\ref{sec:clustering}, the target bucket size $T$
is motivated by $k$-anonymity~\cite{sweeney2002kanonymity}: each proximity
bucket is intended to aggregate $\sim$1{,}000 users, making it difficult to isolate
any individual's contribution to the feature vector.

\paragraph{Geo-IP in industry.}
IP-based geolocation has been used for coarse personalization (e.g., content
localization, ad targeting). Prior work on IP geolocation focuses on the
accuracy and reliability of country/city-level database
mappings~\cite{poese2011ip}, rather than using geo-IP as a basis for
aggregating user-level features for ML. Proximity Features extend this line
of work by operating at sub-city granularity, dynamically adapting resolution
to local user density, and computing rich aggregated user-level features over
geographic groups---purpose-built for ML model serving in a two-sided
marketplace at production scale.

\section{Conclusion}

This paper presented Proximity Features, a privacy-compliant feature system for
cold-start personalization in two-sided marketplaces. The system uses an
adaptive geo-IP clustering algorithm to group users into proximity buckets
targeting approximately 1{,}000 individuals, computes aggregated user-level
features over these groups, and serves them in real time as a soft dependency
for downstream ML models.

Deployed at Airbnb across marketing landing pages and homepage destination
recommendation (AutoSuggest), with serving infrastructure for engagement emails already in place and experiment integration under way, the system
produces statistically significant booking lifts on each online surface. The strongest
gains are among Never-Booked and Dormant users---the population where
traditional user-level features provide the least signal. The system operates
entirely on aggregated, consent-gated data and requires no persistent
individual identifier at inference time, making it resilient to increasing
restrictions on third-party cookies and designed to operate within evolving
privacy regulations.

Several limitations are worth noting. In sparse regions, the coarsening
algorithm produces larger buckets at the cost of geographic precision,
which may reduce feature relevance for rural users. In our internal analysis,
geo-IP accuracy is inherently bounded at approximately 80\% city-level
agreement, meaning a
fraction of users receive features from a geographically incorrect bucket.
These limitations are
largely inherited from the underlying geo-IP data source and are
mitigated by the aggregate nature of the features: individual
misassignments have minimal impact on group-level signal quality.

Beyond the surfaces described in this paper, Proximity Features function as a
reusable feature layer applicable to any model that receives user traffic
with cold-start characteristics. From a two-sided marketplace perspective,
improved demand-side personalization directly benefits supply by reducing
supply-demand mismatch for hosts in high-demand areas. Future directions include hierarchical proximity keys---assigning each IP to
multiple keys at different geographic resolutions to provide multi-scale
features for different use cases, versioned key partitions to
gracefully handle natural shifts in the geo-IP database over time, expanding
destination recommendation to additional surfaces such as Nearby Search,
integration with marketing attribution models, and real-time proximity key
updates to capture intra-day traffic patterns.

\bibliographystyle{ACM-Reference-Format}
\bibliography{references}

\end{document}